\documentclass[a4paper]{article}
\usepackage{wrapfig}
\usepackage{float}
\usepackage[pages=all, color=black, position={current page.south}, placement=bottom, scale=1, opacity=1, vshift=5mm]{background}
\SetBgContents{
	\tt 
}      
\usepackage[utf8]{inputenc}
\usepackage{cite}
\usepackage[margin=1in]{geometry} 

\usepackage{amsmath}
\usepackage{amsthm}
\usepackage{amssymb}

\usepackage[utf8]{inputenc}
\usepackage{hyperref}
\hypersetup{
	unicode,
	pdfauthor={Author One, Author Two, Author Three},
	pdftitle={A simple article template},
	pdfsubject={A simple article template},
	pdfkeywords={article, template, simple},
	pdfproducer={LaTeX},
	pdfcreator={pdflatex}
}

\usepackage[sort&compress,numbers,square]{natbib}

\theoremstyle{definition}

\usepackage{graphicx, color}
\graphicspath{{fig/}}

\usepackage{algorithm, algpseudocode} 
\usepackage{mathrsfs} 

\usepackage{lipsum}

\title{AI-Driven Optimization of Hardware Overlay Configurations}
\author{Rasha Karakchi}

\date{
	University of South Carolina\\ 
    \texttt{karakchi@cec.sc.edu}\\%
	[2ex]%
}

\begin{document}
	\maketitle
	
	\begin{abstract}
Designing and optimizing FPGA overlays is a complex and time-consuming process, often requiring multiple trial-and-error iterations to determine a suitable configuration. This paper presents an AI-driven approach to optimizing FPGA overlay configurations, specifically focusing on the NAPOLY+ automata processor implemented on the ZCU104 FPGA. By leveraging machine learning techniques, particularly Random Forest regression, we predict the feasibility and efficiency of different configurations before hardware compilation. Our method significantly reduces the number of required iterations by estimating resource utilization, including logical elements, distributed memory, and fanout, based on historical design data. Experimental results demonstrate that our model achieves high prediction accuracy, closely matching actual resource usage while accelerating the design process. This approach has the potential to generalize across different FPGA architectures, offering a more efficient workflow for hardware designers.
\end{abstract}
\section{Introduction}
Designing complex systems on Field Programmable Gate Arrays (FPGAs) often presents significant challenges, particularly when using traditional FPGA CAD tools such as Quartus and Xilinx Vivado. One of the most time-consuming aspects of FPGA design is the process of building and compiling the design. This challenge becomes even more pronounced when working with overlays, where the goal is to test whether the hardware resources consumed by the overlay configuration will fit the available FPGA architecture. In these cases, designers are often left to rely on a trial-and-error process, running multiple configurations until they find a solution that works.

One particularly challenging scenario arises when working with specialized FPGA overlays, such as the NAPOLY+ automata processor \cite{karbowniczak2025optimizing, karakc2024overlay, karbowniczak2024scored, karakchi2024scratchpad}. The success of an overlay design depends heavily on the logical features of the dataset, including the number of states and edges in the automata. Each state typically corresponds to a processor element (PE), while each edge corresponds to a wire or hardware interconnect. Theoretically, the number of states and edges should match the number of processor elements and wires, respectively. However, due to factors such as wire congestion and routing limitations, the design may fail to fit, even if these numbers initially appear to align.

In such cases, the designer must adjust the configuration by changing the number of processor elements or interconnections, which may involve extensive experimentation. This trial-and-error approach can result in wasted time, as the designer waits for each configuration to be compiled and tested, only to find that it does not fit the target FPGA architecture. In other instances, while a configuration may fit, the utilization of hardware resources might be low enough that a higher-capacity configuration could have been tested instead, leading to further inefficiency.

The question then arises: How can we reduce the time wasted on these iterative configurations? Traditional methods of designing FPGA overlays are slow, and each compilation cycle can take hours or even days. However, with the integration of artificial intelligence (AI) and machine learning (ML) techniques, there is a promising path forward.

\textbf{Harnessing AI for FPGA Design Optimization:} AI has revolutionized numerous fields by providing solutions to problems that once seemed insurmountable due to their complexity or time-consuming nature \cite{lecun2015deep, silver2016mastering}. In the context of FPGA design, machine learning algorithms, such as the random forest regression, are emerging as powerful tools to predict the likelihood of a given configuration fitting in the FPGA overlay \cite{goswami2021congestion}.

By training a regressor model on historical design data, we can enable the system to predict which configurations are most likely to succeed before the compilation process even begins \cite{bengio2010neural}. For example, in the case of the NAPOLY+ overlay design, the AI model would analyze the number of states and edges in the automata, along with other design parameters, to predict whether the configuration will fit the FPGA hardware. This can dramatically reduce the number of trial configurations that need to be tested manually, saving designers days or even weeks of time spent on tedious trials.

Although the current approach targets a specific overlay (NAPOLY+) and a specific FPGA board (ZCU104), this limitation does not diminish its potential. In fact, this narrow focus allows us to refine and optimize the model for more accurate predictions. The real advantage of this AI-based approach lies in its ability to provide predictions in advance, allowing designers to make informed decisions before committing to lengthy compilation cycles \cite{shi2020ftdl, elsaid2024optimized, noronha2019overlay, karakc2024overlay, Rasha1, Rasha2}.

This work represents just the beginning of what could become a more standardized approach to optimizing FPGA design processes across a range of overlays and FPGA boards. Although the current model is tailored to a specific overlay and FPGA board combination, the ultimate goal is to expand this framework to accommodate a larger variety of FPGA configurations, making AI-driven optimization a tool accessible to all FPGA designers. By leveraging the power of machine learning, we can create more generalizable solutions that will streamline the design process for a wide array of applications.

\section{Targeted Design}
The NAPOLY+ design is a two-dimensional array of specialized STE+ elements, each integrating symbol processing, scoring (weighting) and arithmetic components. These elements accumulate scores along computational paths to determine the final results. The array layout resembles the structure of Configurable Logic Blocks in FPGAs.

Each STE+ process activated signals from multiple predecessors. Unless the "start" bit is set, its state bit resets. When activated, an STE+ transmits an activation signal to its outputs, which is ANDed with an interconnect configuration bit before verifying successors and continuing the process.

Beyond logic activation, STE+ also computes an outgoing score. This score is determined by the incoming score from its predecessor and an edge score, a predefined value stored in a register and configured during reconfiguration. To handle mismatches and gaps, all STE+ elements are connected to the start STE+, though this setup is constrained by array size, fan-out limits, and hardware resources. The interconnection system consists of a grid of global and local wires, with horizontal wires restricted by FPGA bus size (1 million wires) and vertical wires managing local fan-out connections.

During operation, the start STE+ remains active, enabling parallel symbol processing and allowing multiple STE+ elements to function simultaneously. 

To distinguish between new symbols and mismatching inputs, incoming scores are reset to zero for each new symbol, signaling the start of a new path. A dedicated fan-in signal is assigned to new symbols. Accepting STE+ elements confirm matches and report scores upon activation but do not connect to the start state.

In previous work, the performance of NAPOLY+ was evaluated on two Zynq UltraScale+ MPSoC FPGA devices, which are optimized for high-performance applications such as ZCU104 which features approxiately 504K Logical Cells (LCs), 461K Flip-Flops (FFs), and 6.2 Mb of distributed memory. 

\section{Evaluation}
In the context of NAPOLY+ performance prediction, we used the Random Forest Aggressor to estimate the performance efficiency based on different array sizes taken from previous work. This model learns from existing FPGA performance data and generates predictions for new array sizes by analyizing patterns and trends in resource utilization and computational efficiency. 

For training, the model is trained using performance data from ZCU104 NAPOLY+ \cite{karbowniczak2025optimizing, karbowniczak2024scored} which includes array sizes (1K, 2K, 4K, etc.) and their corresponding performance percentages. Multiple decision trees are created, each trained on different subsets of the dataset to ensure robustness. 

For prediction, when given a new array size, the input is passed through all trained decision trees in the random forest. Each tree provides an independent prediction of ZCU104 NAPOLY+ performance based on the learned patterns. The final predicted value is obtained by averaging the outputs from all decision trees, ensuring a stable and accurate estimate. 

\section{Results}
\begin{figure}[]
    \centering
    \includegraphics[width=0.75\linewidth]{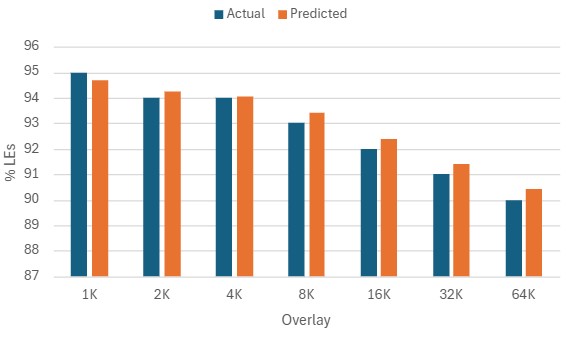}
    \caption{Predicted logical resources versus actual resources over different overlays}
    \label{fig:1}
\end{figure}

\begin{figure}[]
    \centering
    \includegraphics[width=0.75\linewidth]{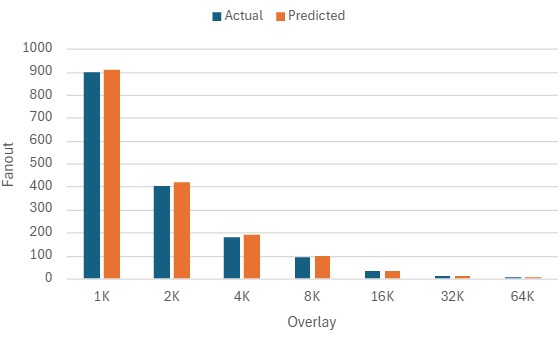}
    \caption{Predicted hardware fanout versus actual resources over different overlays}
    \label{fig:3}
\end{figure}

The three figures below compare the predicted resources with the actual values reported in \cite{karbowniczak2025optimizing, karbowniczak2024scored}. Figure \ref{fig:1} displays the predicted logical resources alongside the actual resources. Figure \ref{fig:3} presents the predicted hardware registers compared to the actual results. Figure \ref{fig:2} presents the predicted distributed memory compared to the actual memory results. Figure \ref{fig:3} illustrates the predicted fanout versus the actual results.

\begin{figure}[]
    \centering
    \includegraphics[width=0.75\linewidth]{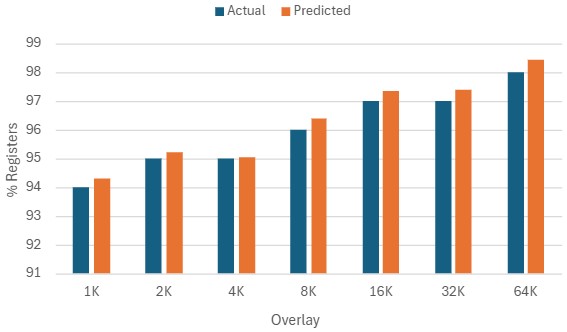}
    \caption{Predicted registers versus actual resources over different overlays}
    \label{fig:2}
\end{figure} 

\begin{figure}[h]
    \centering
    \includegraphics[width=0.75\linewidth]{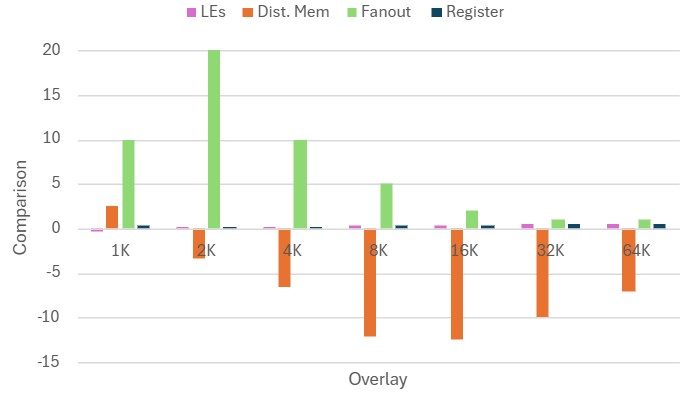}
    \caption{Comparison between the predicted resources and the actual resources}
    \label{fig:4}
\end{figure}

Figure \ref{fig:4} provides a comparison of the results. As observed, the trained model's predictions for the logical resources are the closest to the actual values (similarly the register resutls), evidenced by the smaller discrepancies in the results. In contrast, the predicted memory values were significantly lower than the actual results, as shown by the orange bars in the figure. For the fanout, the trained model slightly overestimated the fanout compared to the actual values.

\begin{figure}[h]
    \centering
    \includegraphics[width=0.75\linewidth]{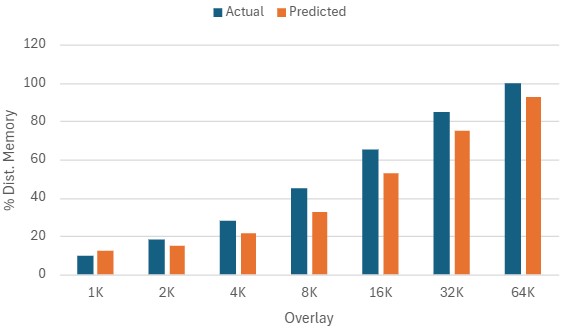}
    \caption{Predicted distributed memory versus actual memories over different overlays}
    \label{fig:2}
\end{figure} 

\section{Conclusion and Future Work}
The primary objective of this work is to reduce the number of attempts required for a designer to identify the configuration set that can be successfully implemented on specific hardware, in this case, NAPOLY+. As an initial step, we employed the Random Forest algorithm to predict whether the results align with the actual outcomes and assess how closely the predictions approximate the actual values.

Since NAPOLY+ was designed specifically for FPGA devices and evaluated on Ultrascale FPGA devices (ZCU104), its performance was assessed in terms of hardware resources consumed and the allowable fanout per configuration. Our model was trained using these performance metrics, considering logical elements, distributed memory, and fanout.

This represents the first step in our approach, with future plans to enhance the model so that it not only predicts resource usage but also determines whether the predicted combination of resources would allow the configuration to fit on the target devices. This advancement will significantly reduce the time required to test various combinations, overlay sizes (represented by the number of STE+), and fanout (hardware interconnections) at maximum frequencies to determine compatibility with the devices. By predicting feasible configurations, the model will help designers avoid much of the trial-and-error process involved in finding the optimal configuration.
\bibliographystyle{plain}

\bibliography{main}

\begin{thebibliography}{10}

\bibitem{karbowniczak2025optimizing}
Ryan Karbowniczak and Rasha Karakchi.
\newblock Optimizing fpga overlays for automata processing.
\newblock {\em Journal of FPGA Research}, 10(1):1--10, 2025.

\bibitem{karakc2024overlay}
Rasha Karakchi.
\newblock {\em An Overlay Architecture for Pattern Matching}.
\newblock PhD thesis, University of South Carolina, 2020.

\bibitem{karbowniczak2024scored}
Ryan Karbowniczak and Rasha Karakchi.
\newblock A scored non-deterministic finite automata processor for sequence alignment.
\newblock {\em arXiv preprint arXiv:2410.19758}, 2024.

\bibitem{karakchi2024scratchpad}
Rasha Karakchi.
\newblock A scratchpad spiking neural network accelerator.
\newblock In {\em 2024 IEEE 3rd International Conference on Computing and Machine Intelligence (ICMI)}, pages 1--5. IEEE, 2024.

\bibitem{lecun2015deep}
Yann LeCun, Yoshua Bengio, and Geoffrey Hinton.
\newblock Deep learning.
\newblock {\em nature}, 521(7553):436--444, 2015.

\bibitem{silver2016mastering}
David Silver, Aja Huang, Chris~J Maddison, Arthur Guez, Laurent Sifre, George Van Den~Driessche, Julian Schrittwieser, Ioannis Antonoglou, Veda Panneershelvam, Marc Lanctot, et~al.
\newblock Mastering the game of go with deep neural networks and tree search.
\newblock {\em nature}, 529(7587):484--489, 2016.

\bibitem{goswami2021congestion}
Pingakshya Goswami and Dinesh Bhatia.
\newblock Congestion prediction in fpga using regression based learning methods.
\newblock {\em Electronics}, 10(16):1995, 2021.

\bibitem{bengio2010neural}
Yoshua Bengio, R{\'e}jean Ducharme, Pascal Vincent, and Christian Jauvin.
\newblock A neural probabilistic language model. journal of machine learning research, 3 (feb): 1137-1155, 2003.
\newblock {\em Google Scholar Google Scholar Digital Library Digital Library}, 2010.

\bibitem{shi2020ftdl}
Runbin Shi, Yuhao Ding, Xuechao Wei, He~Li, Hang Liu, Hayden K-H So, and Caiwen Ding.
\newblock Ftdl: a tailored fpga-overlay for deep learning with high scalability.
\newblock In {\em 2020 57th ACM/IEEE Design Automation Conference (DAC)}, pages 1--6. IEEE, 2020.

\bibitem{elsaid2024optimized}
Kareem Elsaid, M~Watheq El-Kharashi, and Mona Safar.
\newblock An optimized fpga architecture for machine learning applications.
\newblock {\em AEU-International Journal of Electronics and Communications}, 174:155011, 2024.

\bibitem{noronha2019overlay}
Daniel~Holanda Noronha, Ruizhe Zhao, Zhiqiang Que, Jeffrey Goeders, Wayne Luk, and Steve Wilton.
\newblock An overlay for rapid fpga debug of machine learning applications.
\newblock In {\em 2019 International Conference on Field-Programmable Technology (ICFPT)}, pages 135--143. IEEE, 2019.

\bibitem{Rasha1}
Rasha Karakchi, Lothrop~O. Richards, and Jason~D. Bakos.
\newblock A dynamically reconfigurable automata processor overlay.
\newblock In {\em 2017 International Conference on ReConFigurable Computing and FPGAs (ReConFig)}, pages 1--8, 2017.

\bibitem{Rasha2}
Rasha Karakchi, Charles Daniels, and Jason Bakos.
\newblock An overlay architecture for pattern matching.
\newblock In {\em 2019 IEEE 30th International Conference on Application-specific Systems, Architectures and Processors (ASAP)}, volume 2160-052X, pages 165--172, 2019.

\end{thebibliography}
\end{document}